# Improved YOLOv5 Based on Attention Mechanism and FasterNet for Foreign Object Detection on Railway and Airway tracks.

**Zongqing Qi[1*], Danqing Ma[2], Jingyu Xu[3], Ao Xiang[4], Hedi Qu[5]**

[1]Computer Science, Stevens Institute of Technology, Hoboken NJ, U.S,
[2]Computer Science, Stevens Institute of Technology, Hoboken NJ, U.S,
[3]Computer Information Technology, Northern Arizona University, Arizona, U.S,
[4]Digital Media Technology, University of Electronic Science and Technology of China, Sichuan, China.
[5]Computer Science, Shenzhen SmartChip Microelectronics Technology Co., Ltd., China.

[*]The corresponding author Email: zongqingqi@gmail.com.

**Abstract.** In recent years, there have been frequent incidents of foreign objects intruding into railway and Airport runways. These objects can include pedestrians, vehicles, animals, and debris. This paper introduces an improved YOLOv5 architecture incorporating FasterNet and attention mechanisms to enhance the detection of foreign objects on railways and Airport runways. This study proposes a new dataset, AARFOD (Aero and Rail Foreign Object Detection), which combines two public datasets for detecting foreign objects in aviation and railway systems. The dataset aims to improve the recognition capabilities of foreign object targets. Experimental results on this large dataset have demonstrated significant performance improvements of the proposed model over the baseline YOLOv5 model, reducing computational requirements. improved YOLO model shows a significant improvement in precision by 1.2%, recall rate by 1.0%, and mAP@.5 by 0.6%, while mAP@.5-.95 remained unchanged. The parameters were reduced by approximately 25.12%, and GFLOPs were reduced by about 10.63%. In the ablation experiment, it is found that the FasterNet module can significantly reduce the number of parameters of the model, and the reference of the attention mechanism can slow down the performance loss caused by lightweight.

**Keywords:** YOLOv5, FasterNet, NAM, foreign object detection.

## 1. Introduction

As the transportation system continues to evolve, safety concerns regarding transport infrastructure, such as railways and airways, have become increasingly significant. One of the main threats to railway and airways safety is track invasions by obstacles. Common unauthorized rail intrusions include pedestrians, vehicles, animals, and falling rocks. Failure to promptly alert authorities of these intrusions can adversely affect railway traffic safety and operational continuity [1]. This issue is also present in the aviation industry. Foreign debris on airport runways poses a threat to aircraft during takeoff and landing, and in serious cases, can cause irreparable damage [2]. Early methods of protecting against foreign object intrusion on transport routes mainly relied on physical barriers and manual inspections. However, this approach was costly in terms of manpower and finances, and negligence or lax inspections could lead to severe consequences. Traditional, relatively effective detection methods have been based on sensor technology, utilizing various devices such as millimeter-wave radar, and optical cameras to monitor invasions on railways and airport runways [3][4].

In recent years, analysis based on artificial intelligence has developed rapidly in various fields [5][6]. With the development of deep learning and the maturity of video processing technology, detection methods based on computer vision have emerged quickly, reducing a large amount of personnel and financial resources [7][8]. The field of foreign object detection is no exception. However, there is still room for improvement in terms of accuracy and speed when compared to previous detection algorithms. Currently, detection and detection networks are divided into two types: single-stage and two-stage networks [9]. Two-stage networks handle positioning and classification separately, resulting in higher detection accuracy. On the other hand, single-stage networks unify the two, resulting in a significant breakthrough in detection speed compared to the two-stage method. The representative two-stage network is SSD [10] and YOLO [11]. YOLO v5, as one of the mainstream networks currently widely used, performs well in detection accuracy and detection speed. Currently, there are still challenges to be solved when this network is directly applied to the detection of foreign objects in transportation systems. Firstly, railways and airport runways are situated in complex outdoor environments, where factors such as lighting can interfere with identification work. Secondly, foreign objects can have varying shapes in different environments and scenes, which makes feature extraction more difficult. Additionally, computer vision technology often captures images where the identification target is too small, requiring higher precision algorithms to detect them.

To enhance the algorithm's foreign object recognition capabilities, we combined the railway and aviation foreign object detection datasets for training. Additionally, we proposed a new improved YOLOv5 architecture based on FasterNet and attention mechanism to improve the detection ability of traffic runway types.

## 2. Related work

In the structure of the visual field, there are two major categories of methods that are the most popular: one is CNN. CNN is the mainstream architecture sin the field of computer vision, especially in actual deployments that require lightweight design while maintaining a certain level of performance. Specifically, designs such as depth wise separable convolution (or called DWConv) can reduce the parameters and calculation amount of the network structure, such as MobileNets [12]. The other type is a transformer, which represented by ViT [12]. This is a structure based on the attention mechanism derived from the field of natural language processing. Vit exploits a sequence of partial glances and selectivity in focusing on salient parts to better capture visual structure. Although there is a trend to reduce the complexity of the attention operator by using the attention mechanism, it is important to note that attention-based mechanisms generally run slower than convolutional mechanisms, making them less suitable for lightweight development [13].

## 3. Methodology

This section will introduce three network structure changes based on lightweight yolov5.

### 3.1 FasterNet and PConv

To improve yolov5, we first introduced the FasterNet module [14]. The structure of the module and PConv(Partial convolutional layer) is shown in the figure 1.Each FasterNet block has one PConv layer, then two PWConv (PointWise Convolution) layers to increase the number of feature channels and to help feature fusion, and each PConv is followed by two PWConv layers. Together, they are represented as inverted residual blocks, where batch normalization (BN) is chosen to improve the model's training stability and inference speed. PConv is a partially convolutional layer.

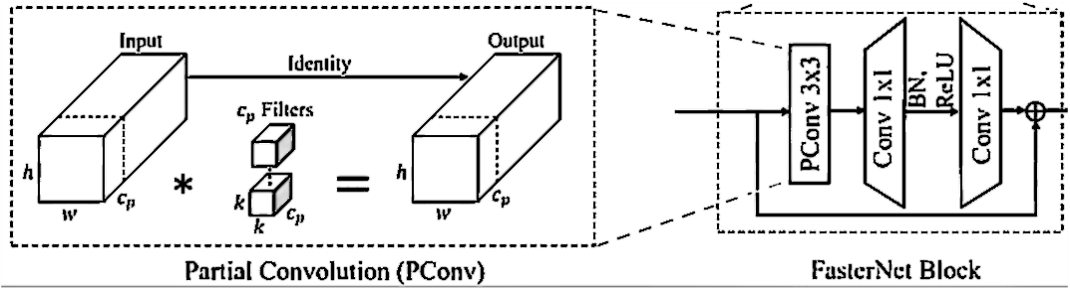

**Figure 1**. FasterNet [14]

In the traditional YOLOv5 convolution operation, the convolution kernel is applied to all channels of the input feature map. However, PConv only selects a continuous subset of channels, either the first or last channels, for convolution operation to extract spatial features. For simplicity, we assume that the input feature map $C_{in}$ and output feature map $C_{out}$ have the same number of channels. The calculation formula for FLOPs is approximately shown in the following equation (1).

$$\text{FLOPs} = C_{in} \times C_{out} \times K^2 \times H \times W \tag{1}$$

Where $K^2$ represents the convolution kernel size，$H \times W$ is the size of the output feature map. It can be easily speculated that the reduction in FLOPs for a PConv compared to a regular convolution is proportional to the square of the partial ratio r. The partial ratio r is expressed as equation (2)：

$$r = \frac{c}{c_p} \tag{2}$$

Where $c$ is the total number of input channels, and $c_p$ is the number of channels participating in convolution. This structural design significantly reduces computing requirements. Although we only use the cp channel for spatial feature extraction, the remaining channels do not need to be removed from the feature map. This allows feature information to flow through all channels, allowing subsequent PWConv layers to obtain complete information, and will not make subsequent Convolutional layers with attention mechanisms useless.

*3.2 NAM*

For the neck network part, this article uses a lightweight attention mechanism module called NAM [15], which uses a serial module similar to CBAM [16]. The integration process of CBAM is shown in Figure 2. NAM redesigns channel and spatial attention sub-modules, embedded at the end of each network block.

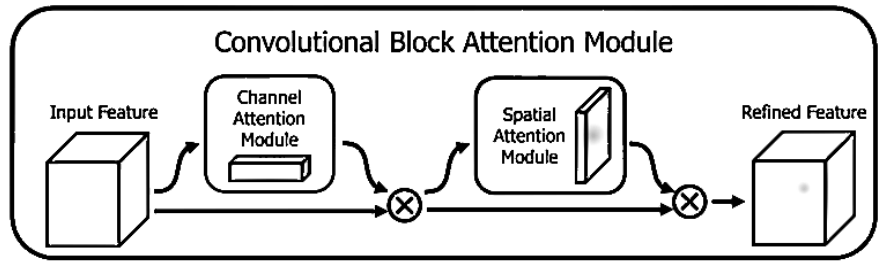

**Figure 2**. Serial module of CBAM [ 16]

It uses a scaling factor for batch normalization that represents the importance of weights through the standard deviation. A scaling factor from batch normalization is used for both channel and spatial attention submodule, which as shown in Equation (3).

$$C_{out} = BN(C_{in}) = \gamma \frac{C_{in} - \mu_C}{\sqrt{\sigma_C^2 + \epsilon}} + \beta \tag{3}$$

Where $C_{in}$ and $C_{out}$ represent the input features and output features; $\mu_C$ and $\sigma_C^2$ respectively denote the mean and variance of the features in the mini batch; $\gamma$ and $\beta$ are adjustable affine transformation parameters used for scaling and shifting the features, $\epsilon$ is a small constant added for numerical stability. The structure of channel and spatial attention submodule is shown in Figure 3. For channel attention submodule, the scaling factor γ measures the importance of each channel, and the weights (Wγ) are obtained by calculating the sum of all channel scaling factors (γj). Similarly, for spatial attention submodule λ is the scaling factor used to measure the importance of pixels in the spatial dimension, and the weights (Wλ) are obtained by calculating the sum of all spatial scaling factors (λj). The final network structure is shown in Figure 4.

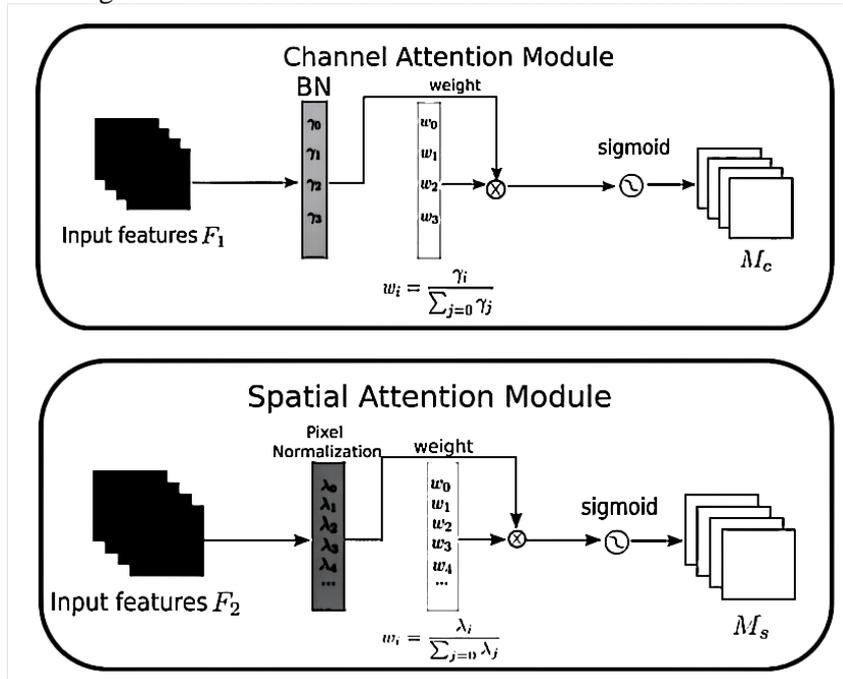

**Figure 3**. The Structure of Channel and Spatial Attention Submodule [15]

*3.2 Dataset*

We present the AARFOD (Aero and Rail Foreign Object Detection) dataset, which is a new large-scale detection dataset specifically designed for foreign object detection in airport runway systems and railway track systems. The AARFOD dataset integrates two public datasets, RailFOD23 [17] and FOD-A [18], comprising a total of 48,409 high-resolution images, covering 35 object categories, and 74,334 annotated objects. The provided images display foreign objects that may cause harm to airplanes and trains, including metal tools, floating objects, bird nests, parts, rubber plastic debris, and branches and leaves found on airport runways and railway tracks. The images were captured under different lighting and weather conditions to simulate real environments and were taken from various distances and angles. To enhance the universality of the dataset and address the scarcity of foreign object images, data augmentation techniques were employed during the construction of AARFOD. These techniques included manual synthesis, automatic generation, background fusion, and noise addition.

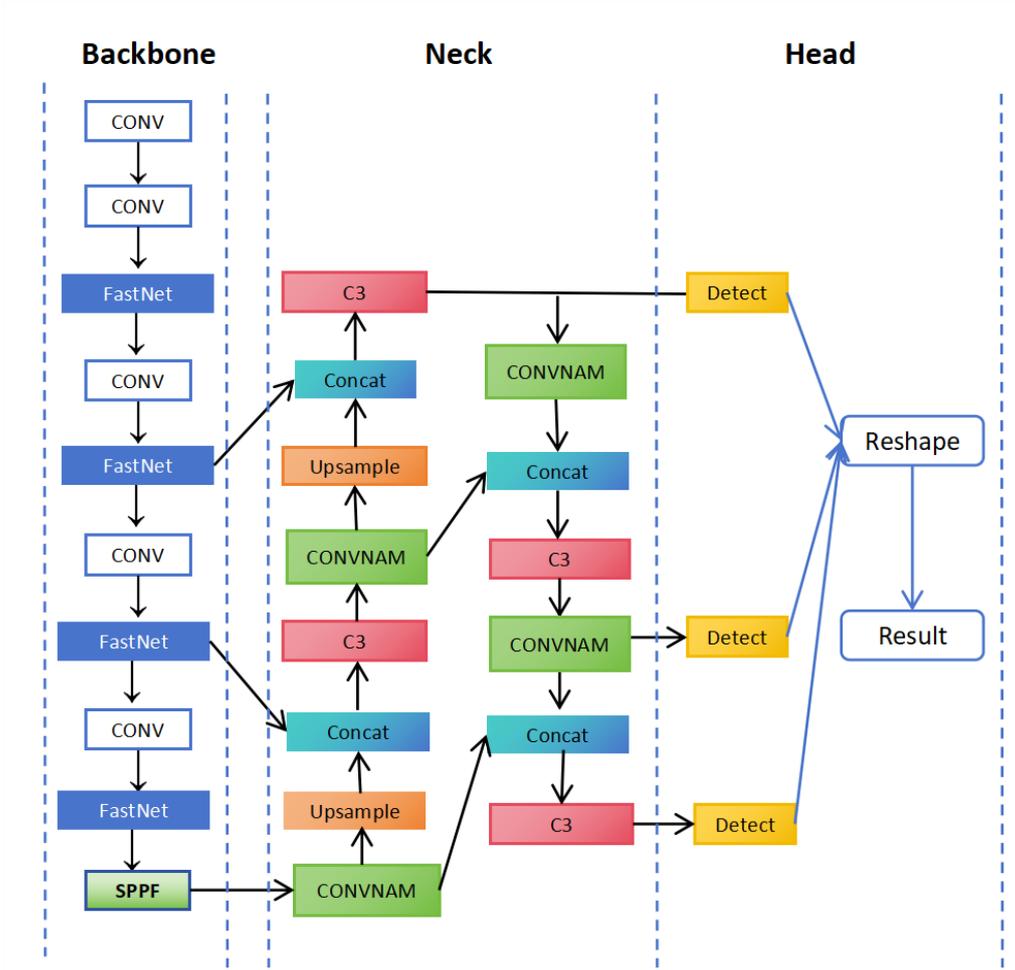

**Figure 4**. The Structure of Final Network structure

## 4. Experiment

The model employs yolov5s as the evaluation benchmark, and all module enhancements are based on yolov5s. The operating system of the machine trained in this experimental model is Windows 11, the CPU model is Intel (R) Core (TM) i3-12400F, the memory size is 32 GB, the GPU model is NVIDIA GeForce GTX 3060Ti, and the graphics card memory size is 8 GB. The evaluation indicators adopt the commonly used precision (P), recall (R), average precision (AP), and mean average precision (mAP) of all categories of AP values in target detection. The calculation formula for the above indicators is as follows:

$$P = \frac{TP}{TP + FP} \times 100\% \tag{4}$$

$$R = \frac{TP}{TP + FN} \times 100\% \tag{5}$$

$$AP = \sum_n P_n \tag{6}$$

$$mAP = \frac{1}{N} \sum_{i=1}^{m} AP_i \tag{7}$$

Precision is the proportion of true positives among the predicted positives, while Recall is the proportion of true positives among all actual positives. Model complexity indicators include the number of parameters and Giga Floating Point Operations (GFLPOs).

Table 1, Figure 5 and Figure 6 show the experimental result data, where YOLO-FasterNet means that FasterNet is replaced only on the head, and YOLO-NAM means that the attention mechanism module NAM is loaded on the neck. The results show that all models have relatively high precision and recall, especially the mAP@.5 values of all models are close to or above 0.98. This is due to the large and merged dataset, which can make a strong contribution to foreign object recognition training support.

**Table 1.** Model training time results

| Model | Precision | Recall | mAP@.5 | mAP@.5-.95 | Parameters(M) | GFLOPs |
|-------|-----------|--------|--------|------------|---------------|--------|
| YOLOv5s | 0.967 | 0.971 | 0.982 | 0.873 | 7.102 | 14.822 |
| YOLO-FasterNet | 0.974 | 0.979 | 0.987 | 0.868 | 5.212 | 12.181 |
| YOLO-NAM | 0.971 | 0.977 | 0.986 | 0.874 | 7.282 | 16.124 |
| YOLO-Improved | 0.979 | 0.981 | 0.988 | 0.874 | 5.318 | 13.246 |

Compared to the baseline model, YOLO-FasterNet shows an increase in accuracy by 0.7%, recall rate by 0.8%, and mAP@.5 by 0.5%. However, there is a slight decrease of 0.5% in mAP@.5-.95. The number of parameters reduces by approximately 26.61%, and GFLOPs decrease by approximately 17.82%. Similarly, YOLO-NAM exhibits an increase in accuracy by 0.4%, recall rate by 0.6%, and mAP@.5 by 0.4%. There is also a slight increase of 0.1% in mAP@.5-.95. However, the number of parameters increases by approximately 2.53%, and GFLOPs increase by about 8.78%. In comparison to the baseline model, improved YOLO significantly improves in precision by 1.2%, recall rate by 1.0%, and mAP@.5 by 0.6%, while mAP@.5-.95 remains unchanged. The number of parameters reduces by approximately 25.12%, and GFLOPs decrease by about 10.63%.

The experimental results indicate that YOLO-FasterNet slightly improves the overall detection performance while maintaining similar mAP@.5-.95 performance, and significantly reduces the model's parameter size and computational requirements. This demonstrates that the module design of FasterNet is indeed optimized for speed. Similarly, YOLO-NAM also improves the overall performance, particularly in terms of precision and recall rate, with a slight increase in computational load. Introducing a normalized attention mechanism effectively focuses on key feature channels. By integrating the two approaches, YOLO-Improved significantly enhances recognition performance while reducing parameter size and computational complexity. The model achieves great results through innovative architectural improvements. As shown in Figure 5, the model achieves high accuracy in object recognition, which has important implications for its deployment on real-world lightweight devices.

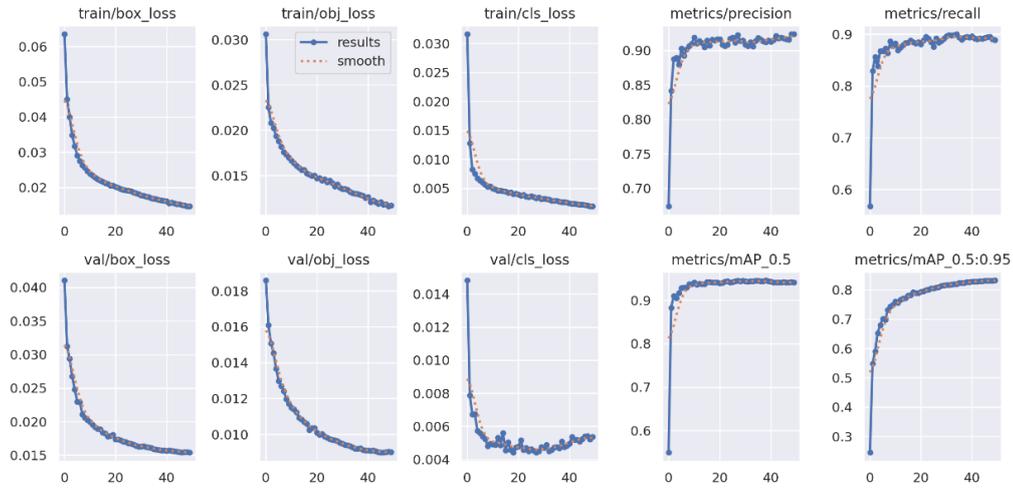

**Figure 4**. The training process of Fostc3net

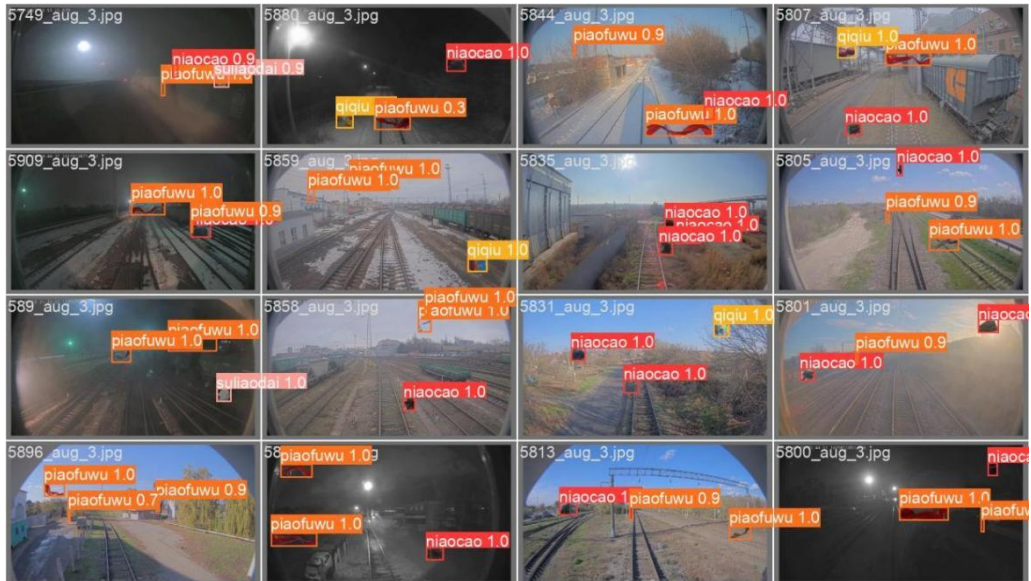

**Figure 5.** The actual effect after Fostc3net training

## 5. Conclusion

This paper discusses the issue of foreign object intrusion in transportation infrastructure, specifically targeting railways and airport runways. To address this challenge, this paper introduces an improved YOLOv5 architecture that combines the NAM module from FasterNet with an attention mechanism to enhance the understanding of transportation environments and detect foreign objects on runways. The improvement to the YOLOv5 architecture involves utilizing Partial Convolutions (PConv) for efficient feature extraction and a lightweight attention mechanism module, NAM, for enhanced feature recognition. Additionally, the article proposes a new dataset named AARFOD. This dataset combines two public datasets to support the training and evaluation of the proposed detection system under various conditions. The experimental results demonstrate that the proposed improvements significantly enhance precision, recall, and mean average precision (mAP) compared to the baseline YOLOv5 model, while also reducing computational requirements. Future research may delve deeper into the structure, particularly in exploring more efficient ways to combine FasterNet and attention mechanisms for more effective recognition outcomes.